\documentclass[journal,table]{IEEEtran}

\usepackage{todonotes}
\usepackage{wrapfig}
\usepackage{dblfloatfix}
\usepackage{subfig}

\usepackage{tcolorbox}     
\usepackage{float}         
\usepackage{lipsum}        

\usepackage{cite}
\usepackage{enumitem}

\floatstyle{plain}
\newfloat{highlightbox}{t}{lop}  
\floatname{highlightbox}{Note}   

\newcommand{\GreenPlusCircle}[1][0.15cm]{%
    \tikz[baseline=(plus.base)] {
        \fill[green] (0,0) circle (#1);
        \node[font=\bfseries,scale=0.8] (plus) {+};
    }%
}

\newcommand{\EmptyCircle}[1][0.15cm]{%
    \tikz[baseline=(plus.base)] {
        \draw[fill=white, draw=black] (0,0) circle (#1);
        \node (circle) {};
    }%
}

\newcounter{mybox}

\ifCLASSINFOpdf
\else
\fi
\begin{document}
\title{{Robot-mediated physical Human-Human Interaction in Rehabilitation: a position paper}}

\author{Lorenzo Vianello, Matthew Short, Julia Manczurowsky, Emek Barış Küçüktabak, Francesco Di Tommaso, \\ Alessia Noccaro, Laura Bandini, Shoshana Clark, Alaina Fiorenza,  Francesca Lunardini, Alberto Canton, \\ Marta Gandolla, Alessandra L. G. Pedrocchi, Emilia Ambrosini, Manuel Murie-Fernández, Carmen B. Román,\\ 
Jesus Tornero, Natacha Leon, Andrew Sawers, Jim Patton, Domenico Formica, Nevio Luigi Tagliamonte,  \\ Georg Rauter, Kilian Baur, Fabian Just, Christopher J. Hasson, Vesna D. Novak, Jose L. Pons 
\thanks{L.V., M.S., E.B.K., S.C., A.F., J.P., and J.L.P. are with Shirley Ryan AbilityLab, Chicago, IL, USA. M.S., E.B.K., J.L.P. are with Northwestern University, Evanston, IL, USA. J.M. and C.J.H. are with Northeastern University, Boston, MA, USA. V.N. is with the University of Cincinnati, Cincinnati, OH, USA. F.D.T. and  N.L.T. are with the Università Campus Bio-Medico di Roma, N.L.T.is with the Fondazione Santa Lucia, Rome, Italy. L.B. is with the University of Genoa, Genoa, Italy. A.S. and J.P. are with the University of Illinois in Chicago, Chicago, IL, USA. M.M-F., C.B.R. are with the Canarian Foundation Institute of Neurological Sciences, Spain. F.J. is with the Chalmers University of Technology, Gothenburg, Sweden and the Institute of Biomedical Engineering, Ulm University, Germany. K.B. is with the Swiss Federal Institute of Technology, Zurich, Switzerland. G.R. is with University of Basel, Basel, Switzerland. M.G., A.L.G.P., and E.A. are with the Politecnico di Milano, Lecco, Italy. J.T., N.L., F.L., and A.C. are with the Hospital Los Madroños, Madrid, Spain, and the Universidad Carlos III de Madrid, Madrid, Spain. A.N. and D.F. are with Newcastle University, UK.}
\thanks{This work was supported by the National Science Foundation~/~National Robotics Initiative (Grant No: 2024488 and 2024813). EA, AP and MG were supported by the Italian Ministry of University and Research within the project HYBR-ID (Project number: 2022F3JPLY; CUP: D53D23001230006; Fundings: Next Generation EU) and by INAIL (Istituto Nazionale Assicurazione contro gli Infortuni sul Lavoro), within the FeatherEXO (PR23-RR-P1) project. Manuscript submitted July 11th, 2025. (correspondence e-mail: lvianello@sralab.org)}
}

\markboth{IEEE Reviews in Biomedical Engineering 1937-3333}%
{Shell \MakeLowercase{\textit{\textit{et al.}}}: Bare Demo of IEEEtran.cls for IEEE Journals}

\maketitle

\begin{abstract} 
Neurorehabilitation conventionally relies on the interaction between a patient and a physical therapist. Robotic systems can improve and enrich the physical feedback provided to patients after neurological injury, but they under-utilize the adaptability and clinical expertise of trained therapists.
In this position paper, we advocate for a novel approach that integrates the therapist’s clinical expertise and nuanced decision-making with the strength, accuracy, and repeatability of robotics: \textit{Robot-mediated physical Human–Human Interaction}. This framework, which enables two individuals to physically interact through robotic devices, has been studied across diverse research groups and has recently emerged as a promising link between conventional manual therapy and rehabilitation robotics, harmonizing the strengths of both approaches.
{Although current findings are largely based on pilot studies and conceptual frameworks, integrating therapists' expertise with the functionalities offered by robotic systems represents a promising direction for improving rehabilitation outcomes.}
This paper presents the rationale of a multidisciplinary team---including engineers, doctors, and physical therapists---for conducting research that utilizes: a unified taxonomy to describe robot-mediated rehabilitation, a framework of interaction based on social psychology, and a technological approach that makes robotic systems seamless facilitators of natural human-human interaction.

\end{abstract}


\IEEEpeerreviewmaketitle

\section{Position}

The common trope regarding robots is that they could gradually replace skilled human jobs. The same concern arises in neurorehabilitation (NR). The use of robots in rehabilitation settings offers exercise automation and reduced therapist effort. However, these benefits are overshadowed by the limited impact they currently have on improving motor learning and incorporating clinician input \cite{celian2021day, tamburella2024quantifying,  yamamoto2022effect}. 
In this position paper, we present a rationale for using robots in NR with the ultimate goal of improving clinical outcomes. To achieve this, we propose leveraging robotics to facilitate therapist-patient physical interaction in a way that preserves the nuanced, skilled care provided by experienced physical therapists.
The integration of these novel technologies, which combine the adaptability, flexibility, and social interaction of physical human-human interaction with the physical strength and controllability of robots, is poised to have a transformative impact on clinical care. {To achieve this long-term goal, statistically powered clinical studies are necessary to validate the expected improvements in functional training, therapy dose, quality of care, and healthcare cost reduction.}


\section{Background}
Millions of people around the world face significant challenges due to sensorimotor impairments associated with neurological conditions such as stroke and spinal cord injury (SCI) \cite{feigin2025world, singh2014global}.
These impairments severely restrict their independence, employment opportunities, and diminish overall quality of life \cite{carod2009quality, post2005quality}, while also imposing a profound economic burden on families and healthcare systems \cite{diop2021quality, strilciuc2021economic, lucas2023costs}.

{Individuals with sensorimotor impairments routinely engage in functional rehabilitation exercises in clinical settings to regain control of their movements, guided by physical therapists who play a central role by interpreting patient states, modulating assistance, and ensuring safety \cite{shumway2007motor, higgs2024clinical}. They design and adapt individualized rehabilitation plans in collaboration with clinicians, tailoring interventions to each patient’s condition and progress. Through empathy and affective perception, therapists provide personalized feedback and adjust therapy based on patient readiness, fatigue, and emotional state \cite{wulf2016optimizing, deci2008self}.} 

Importantly, therapists are trained to use physical contact to assess and treat patients. Through haptic feedback, they evaluate motor abilities, identify deficits, and guide limb movements during exercises.
Physical contact provides both explicit feedback and implicit proprioceptive input, which can complement or compensate for impaired somatosensation. This interaction facilitates the internalization of correct motor patterns \cite{hornby2020clinical} and ensures task completion and reinforcing success during intensive, repetitive training \cite{marchal2009review}.

Helping patients with whole-body movements or severe impairments places a significant physical burden on therapists. While high-intensity, repetitive training is essential for effective rehabilitation, it further adds to this strain \cite{burnfield2013comparative, boukadida2015determinants}. This physical strain increases the risk of musculoskeletal injuries \cite{mccrory2014work} {and can negatively affect task execution, e.g., by reducing smoothness and coordination, and may further hinder the therapist’s ability to effectively supervise the rehabilitation process.} In some cases, optimal care requires multiple therapists, compromising therapist availability. This overall complexity also makes accurate measurement difficult and complicates the characterization and repetition of therapeutic procedures \cite{sawers2014perspectives, galvez2005measuring}.

Rehabilitation robots help address these challenges by providing precise, repetitive practice through assistive or resistive forces \cite{gassert2018rehabilitation}.
They facilitate the assessment of patient performance and allow objective quantification of progress throughout the rehabilitation process. 
However, the use of rehabilitation robots often entails the loss of physical contact between therapists and patients, removing a critical channel for understanding each other's intentions and promoting motor skill acquisition. In this case, a less experienced therapist may retreat into passive supervision and simple parameter tweaks, resulting in an under-utilization of both the robot’s capabilities and their own clinical expertise \cite{celian2021day, postol2023we}.

This paper highlights a novel framework that integrates therapist expertise with the strength and repeatability of robotics through \textit{Robot-mediated physical Human-Human Interaction (RHHI)}.  The paper is structured as follows: Sec. \ref{sec:taxonomy} introduces the salient aspects of RHHI and a taxonomy of this novel framework. Sec. \ref{sec:RmpHHI_Neurorehab} presents how this approach can be used in NR to simultaneously leverage the strengths of therapists and robotic devices. Sec. \ref{sec:applications} outlines examples of potential applications of these technologies in the context of conventional rehabilitation practice. Sec. \ref{sec:KeyConsideration} discusses challenges associated with the implementation of these technologies. Finally, Sec. \ref{sec:open_research_questions} explores open questions and future research directions.


\section{Taxonomy and Overview of RHHI in NR}
\label{sec:taxonomy}

In RHHI, one or more robotic devices are used to render specific interaction dynamics between individuals \cite{Ganesh2014, Takagi2018HapticInteraction, kuccuktabak2021human}.  
In NR, at least one individual is a patient physically connected to a robot, whose movements are influenced in real time by another user, {such as a therapist, another patient, or even a family member}. Together, the individuals engage in coordinated movements aimed at promoting motor learning for the patient(s) and facilitating the transfer of these skills to everyday functional activities.
To better analyze RHHI scenarios in NR, the following section introduces key characterizations.

\subsection{Interaction Configurations}
\label{sec:int_config}

Depending on the type and complexity of the robotic devices used, RHHI enables physical interaction at different levels:

\textit{Task-Space interactions}, where users exchange forces through the end-effector of a robot (Fig. \ref{fig:HRRI}.A and Fig. \ref{fig:HRRI}.G), without specific postural corrections.  \textit{Single-joint interactions}, where the two users interact by moving a single joint (e.g., in Fig. \ref{fig:HRRI}.B, the wrist). This interaction is mediated using a single-DoF robots and allows joint-specific training. \textit{Multi-joint interactions} allows coordination across multiple DoFs (Fig. \ref{fig:HRRI}.C and Fig. \ref{fig:HRRI}.E). This type of interaction is typically achieved with exoskeletons. \textit{Cross-limb interactions}, {in which the involved parts of the bodies of the two persons are not the same, such as the upper body of one user is connected to the lower body of another} (Fig. \ref{fig:HRRI}.G).  \textit{Guided interactions}, where one person interacts with a robot and another person can influence the robot’s behavior, for example, by teaching a kinematic or force profile while directly interacting with the robot (Fig. \ref{fig:HRRI}.D and {discussed in Box.} \ref{box:lbd}).

RHHI can accommodate various interaction configurations, including bidirectional and unidirectional interaction. The interaction is defined as 
\textit{bidirectional} if both users perceive interaction forces due to the interaction between them. {This configuration can be implemented, for example, by either using two robotic devices that both render interaction forces to each user} (Fig. \ref{fig:HRRI}.E) or using a transparent robotic device that directly transmits forces applied by one user to a second user (Fig. \ref{fig:HRRI}.D).
\textit{Unidirectional} interaction refers to a setup in which only one user perceives forces through a robotic device, while the device’s motion is teleoperated or influenced by another user.
This can be achieved in real time (Fig. \ref{fig:HRRI}.E,F) or via prior demonstrations and replication using a single device (Fig. \ref{fig:HRRI}.D), either teleoperated or influenced by another user, or with two similar devices, where only one renders forces.

\subsection{Interaction Medium}
In RHHI, the \textit{interaction medium} refers to the virtual connection implemented between two users through robotic devices. This connection can be designed to render specific mechanical behaviors, such as a virtual spring-damper model between joint positions, that enables users to feel each other’s movements through adjustable mechanical impedance levels~\cite{Takagi2018HapticInteraction, Che2016TwoCoupling, Ganesh2014, Tanaka2019ATask, Piovesan2013HapticTele-interaction}. A rigid virtual spring constant results in closely coupled motion \cite{Takagi2018HapticInteraction, short2023haptic}. In contrast, a compliant, softer virtual spring allows more independent movement while still facilitating the exchange of forces. An asymmetric interaction medium can also be implemented, such as the case where one user experiences a rigid connection while the other experiences a soft one \cite{noccaro2025robot}, enabling customized feedback and workload distribution. It is important to note that the resulting behavior is not solely determined by the virtual model, but also by the human biomechanics, such as the stiffness of the involved joints, residual strength, and active participation of each user. Indeed, individuals can modulate their joint stiffness to either resist or follow the forces transmitted through the interaction \cite{borner2023physically}.

\begin{figure*}[tp]
    \centering
    \includegraphics[width=0.99\linewidth]{imgs/schemaAllTogheter2.png}
    \caption{\small{{Examples of Robot-mediated physical Human-Human Interaction (RHHI) in neurorehabilitation (NR): (A) Two users (typically a patient and a partner such as a therapist, another patient, or a family member, \textit{Partner Characteristics}) interact through two robots. The \textit{Task} definition influences the User strategies and their performance. (B) \textit{Multiple patients} can perform interactive tasks by virtually connecting multiple devices.
    (C) Therapist and patient interact via two upper-limb exoskeletons. They can assume different roles during the interaction (\textit{Interaction Scenario}). (D) Robotic devices can be used for \textit{learning by demonstration} strategies.
    (E-F) The \textit{interaction medium} between the two users is modeled using a virtual spring-damper system characterized by therapist and patient stiffness ($K_T, K_P$) and damping ($D_T, D_P$) during \textit{bi-directional} (E) and \textit{uni-directional} (F) interaction. The interaction between users can be in joint-space (E) or in task-space (G). 
    }}}
    \label{fig:HRRI}
\end{figure*}


\subsection{Partner {Characteristics}} \textit{Partner characteristics} refer to individual and relational factors that influence how users behave during RHHI. These include the relative skill levels of each user, as well as their social relationship (e.g., strangers, friends, family members, peers, or in a patient-therapist dynamic)~\cite{Che2016TwoCoupling, AvilaMireles2017SkillInteraction,Ganesh2014,Kager2019TheTask,Takagi2019IndividualsGoal,Mace2017BalancingTraining,B2018HapticField, Gorsic2017CompetitiveIntensity,Gorsic2017AGame}. Such characteristics can affect muscular effort, response strategies, task engagement, and the roles that emerge during the interaction. For example, a more skilled or confident partner may naturally take on a guiding role, while a less experienced partner may assume a more passive role. {In the case of peer-based interactions, e.g., between two patients, it is particularly valuable to enable the scaling of targets, task difficulty, and interaction roles to address the individualized therapeutic objectives of each participant.}

Another important dimension of the framework is the \textit{number of users} participating in RHHI. Most studies focus on dyadic interactions, in which two individuals are physically connected through one or two robotic devices.  However, multi-user configurations have also been explored \cite{Takagi2019IndividualsGoal}. While these setups present additional challenges in terms of control complexity and task coordination, they offer unique opportunities for social engagement, scalability, and group-based rehabilitation.


\subsection{Interaction Scenario} The \textit{interaction scenario} defines the way the task is structured and how roles are assigned between participants \cite{Jarrasse2012}. Common scenarios include collaboration, cooperation, and competition. These modes of interaction can significantly shape user behavior, influencing engagement, learning strategies, and the physical dynamics of interaction. 
In \textit{collaborative} scenarios, both users share the same task goal and work together without predefined roles \cite{Che2016TwoCoupling}.
For example, a patient may perform a task jointly with a therapist, another patient, or a family member, sharing the effort required to complete movements that might otherwise be too difficult alone.
In \textit{cooperative} scenarios, users still share a goal, but with distinct roles such as teacher/student \cite{Aoyagi2007, Banala2007, Kim2010} or leader/follower \cite{Chackochan2019IncompleteAction}. 
A therapist, for instance, may lead the interaction by providing guidance while the patient adopts a more passive role. {Alternatively, in a home scenario, a family member could guide a patient with the help of a robot \cite{Baur2018RobotSupportedTraining}. Finally, in a clinical setting, a less severely impaired patient could potentially guide and support a more severely impaired patient or could take on a more difficult role in a shared task \cite{GorsicRehab2020}.}
In contrast, \textit{competitive} scenarios involve opposing goals, often implemented through game-based tasks \cite{Gorsic2017AGame}. 
Competitive strategies can be employed to encourage patients' active engagement and augment the challenge of a given task, though this would require careful balancing of exercise difficulty to ensure that, for example, a less impaired patient does not immediately overwhelm a more severely impaired patient \cite{Gorsic2017AGame}.

\subsection{Task Performance and Motor Learning} \textit{Task performance} in RHHI quantifies how users perform the assigned motor task while physically connected. Performance is typically measured through metrics such as accuracy, timing, smoothness, or synchronization. It is influenced by the design of the interaction medium, the characteristics of the users, and the selected interaction scenario. Physical feedback exchanged between users has been shown to improve accuracy, smoothness, coordination and overall task efficiency \cite{Reed2006HapticMachines, Mojtahedi2017OnDyads, Takagi2018HapticInteraction, ivanova2022interaction}. 

\textit{Motor learning} refers to persistent improvements in motor performance due to the RHHI training, after the physical connection is removed. 
This lasting change represents the core therapeutic effect in NR, reflecting the patient's ability to retain and apply improved motor strategies independently.
This is often assessed by comparing individual performance (i.e., without the physical interaction) before and after training. RHHI has been shown to support motor learning through various mechanisms, including increased motivation and engagement \cite{Gorsic2017CompetitiveIntensity, GorsicRehab2020, GajadharDoubleFun}, the opportunity to estimate and internalize a partner’s movement strategy \cite{Takagi2017, Takagi2018HapticInteraction, Takagi2019IndividualsGoal} and role specialization that emerges during interaction \cite{Reed2006HapticMachines, Mojtahedi2017OnDyads}. These benefits make RHHI particularly promising for NR, where one of the users aims to regain lost motor functions through repeated, adaptive practice.

\section{RHHI Potential in NR}
\label{sec:RmpHHI_Neurorehab}

RHHI combines the advantages of robotics (Sec. \ref{sec:RobotAddedValue}) while incorporating {the therapist's expertise and the human adaptability into the control} (Sec. \ref{sec:TherapistExperience}). This section summarizes both contributions (Tab. \ref{tab:conventional_rehab_comparison}).

\subsection{Added Value of Robotics}
\label{sec:RobotAddedValue}
Robotic devices complement therapists by partially offloading the patient's weight and accurately transmitting the therapist's intended movements to the patient \cite{gassert2018rehabilitation}. This feature is particularly valuable during prolonged and intensive rehabilitation sessions that require continuous \textit{physical effort}. It is especially beneficial for patients with severe motor impairments, as therapists must maintain patient posture, joint alignment, and smooth movements throughout an exercise. Moreover, the \textit{high repeatability} of robotic devices can relieve therapist effort by replicating movements from previous demonstrations.

In addition, robotic platforms enrich \textit{rehabilitation data} by capturing simultaneous kinematic and dynamic information from both the patient and the therapist \cite{riener2006human}. Kinematic data provide movement trajectories, joint angles, and velocities, revealing patterns and achievable ranges of motion. Dynamic data, such as interaction forces, account for the level of assistance or resistance provided and patients' kinetic responses, giving deeper insight into the quality of the interaction. 

RHHI holds great potential beyond clinical settings. In recent years, the availability of commercially-available assistive devices intended for domestic use has been increasing \cite{sunny2023assistive}. Applying RHHI to these devices could enable continued support for exercises outside of the clinic.
For individuals requiring continuous care, RHHI enables \textit{remote assistance} and rehabilitation, allowing caregivers and therapists to monitor activities, provide real-time support, and potentially ensure a more effective continuum of care, even in situations where physical proximity is limited, such as for patients with autism \cite{santos2021design, fachantidis2020tauhe, jouaiti2019robot}.

\subsection{Added Value of Therapists {and Other Humans}}
\label{sec:TherapistExperience}
In contrast to traditional robotic systems, which typically rely on predefined trajectories and often require time-consuming calibration, {RHHI allows exercise customization and adjustment on-the-fly by incorporating \textit{therapist expertise}, human adaptability, and the ability to observe the patient directly. While this would most obviously be done by therapists, who are trained to support patients, it could (with proper robotic support) be administered by caregivers \cite{Baur2018RobotSupportedTraining} or potentially even less severely impaired patients.}

Moreover, traditional systems often require settings to be adjusted by non-technical users, like clinicians, who may not fully understand their impact, potentially limiting the effectiveness and usability of these systems. 
{Whereas in RHHI, the therapist's intervention dynamically shapes the robotic response, allowing for real-time \textit{adaptation} based on patient feedback and performance, enhancing intuition in robotic interventions}
\cite{hasson2023neurorehabilitation, amato2024unidirectional}. During each session, therapists can flexibly select and modify exercises or movements based on immediate and precise patient assessments without having to set predefined trajectories.

Therapists can also modulate assistance levels throughout rehabilitation. They may initially offer substantial support, provide transparent supervision during assessments, or introduce resistance to encourage active engagement. Although quantifying assistance or resistance can be difficult, therapists can intuitively assess these needs through direct interaction and visual observation, as in conventional therapy \cite{baur2019beam, luciani2024therapists} {and caregivers or even patients themselves can potentially perform the adaptation themselves if a therapist is not available \cite{Gorsic2017ComparisonExercises}.}

Interacting with a human partner further enhances \textit{motivation} and outcomes by harnessing the emotional and psychological benefits of \textit{social engagement}. Patients can interact with therapists, peers, or family members, selecting partners based on therapeutic objectives such as assessment, participation or challenge-based training. The effectiveness of these interactions depends on the availability of the partner, patient preferences, and individual characteristics such as personality type and cognitive ability.

\begin{table}[t]\centering
\begin{tabular}{|c|c|c|}
\hline
Feature & Therapist Added Value & Robot Added Value \\
\hline \hline
Therapist expertise & \GreenPlusCircle[0.12cm] &  \EmptyCircle[0.12cm] \\  
Exercise customization & \GreenPlusCircle[0.12cm] & \EmptyCircle[0.12cm] \\  
Patient adaptability & \GreenPlusCircle[0.12cm] & \EmptyCircle[0.12cm] \\  
Social interaction & \GreenPlusCircle[0.12cm] & \EmptyCircle[0.12cm] \\  
Reduced therapist effort & 
\EmptyCircle[0.12cm] 
& \GreenPlusCircle[0.12cm] \\  
Measurability & 
\EmptyCircle[0.12cm] 
& \GreenPlusCircle[0.12cm] \\  
Repeatability & 
\EmptyCircle[0.12cm] 
& \GreenPlusCircle[0.12cm] \\  
Remote rehabilitation & 
\EmptyCircle[0.12cm] 
& \GreenPlusCircle[0.12cm] \\  
\hline
\end{tabular}
\caption{\small
\small{Added values of therapist and robotic devices in RHHI.}}
\label{tab:conventional_rehab_comparison}
\end{table}

\section{RHHI Applications in NR}
\label{sec:applications}

The potential applications of RHHI in NR are diverse and far-reaching. This section begins by presenting a potential implementation of RHHI for impairment assessment (Sec. \ref{sec:app:assessment}), followed by three different training examples (Sec. \ref{sec:app:UpperLimb}, \ref{sec:app:sitToStand},  \ref{sec:app:gait}). Finally, we explore the use of RHHI in group therapy (Sec. \ref{sec:app:group}) and remote therapy (Sec. \ref{sec:app:remote}). {All approaches are supported by pilot studies; however, more statistically powered investigations are needed to assess their actual effectiveness.}

\subsection{Motor Assessment}
\label{sec:app:assessment}

Assessing patient impairment levels is crucial in rehabilitation \cite{riener2006human, longatelli2023instrumented}. A thorough evaluation must capture four domains: pathology, impairment, participation, and contextual factors \cite{wade2001research}. 
When such information is recorded in a standardized, well-structured format, this enriches insight into the rehabilitation process, streamlines data sharing and reuse, and ultimately delivers more cost- and time-efficient care \cite{berenspohler2021common}. Assessing physiological changes {(Box \ref{box:physSign})} in the brain or spinal cord to understand motor recovery is often not clinically feasible. Therefore, proxies such as muscle weakness, spasticity, and motor coordination are often evaluated and used as markers for recovery. Functional task scales, such as the 9-hole peg test and Box \& Blocks,  offer valuable information about a person’s ability to perform daily activities \cite{stucki2007international}. Because these metrics can change over time, continuous monitoring is crucial to adapt therapy to patient needs.

Combining the quantitative precision of robotic devices with the personalized insights of manual assessments through RHHI enables a comprehensive and adaptive evaluation approach.
An innovative RHHI example is the “Beam-Me-In” strategy 
presented by Baur \textit{et al.} \cite{baur2019beam}. Here, a therapist and patient are connected through exoskeleton robots, enabling therapists to directly perceive patient limitations (e.g., muscle tone, range of motion). This embodied experience provides therapists with a more accurate understanding of patient impairments compared to traditional observation alone. RHHI simultaneously quantifies joint kinematics and interaction forces, enhancing the precision of impairment assessments and treatment adjustments. At the same time, the therapist can design assessment movements in real time, informed by clinically relevant parameters---such as perceived effort, fatigue, or pain---that are difficult to quantify objectively. {Box. \ref{box:physSign} visualize how physiological signals can complete the therapist assessment and allow a better understanding of the patient's impairment during HRRI.}

\begin{highlightbox*}
\refstepcounter{mybox}
\begin{tcolorbox}[colback=blue!5!white, colframe=blue!75!black, title= Box 1: Physiological Signal and RHHI, 
fonttitle=\bfseries, boxsep=2pt,    
  left=4pt,       
  right=4pt,      
  top=2pt,        
  bottom=2pt      
  ]
\label{box:physSign}
Electromyography (EMG) quantifies volitional input during RHHI and reveals how muscles co-contract to stabilize joints and adapt to partner movements and task uncertainty \cite{short2025effects, Takagi2018HapticInteraction, borner2023physically}.
In teleimpedance control \cite{peternel2022after}, EMG provides real-time \textit{\textbf{muscle activity}} data, enabling robots to dynamically adjust impedance. Integrating teleimpedance into RHHI could allow bidirectional control of joint posture and user impedance. Combined with Functional Electrical Stimulation (FES), therapist EMG signals could trigger patient stimulation to promote synchronized movement and motor relearning \cite{dalla2024hybrid}.

Electroencephalography (EEG) can be used to assess and quantify Interpersonal Synchronization, and simultaneous EEG measurements, known as EEG hyperscanning, have been used to provide insights into \textit{\textbf{neural synchronization}} between individuals engaged in shared activities \cite{short2021eeg}. Research has shown that inter-brain synchronization, particularly in regions associated with the mirror neuron system \cite{van2009understanding}, increases during coordinated activities \cite{dumas2010inter, yun2012interpersonal}. 

Autonomic Nervous System response, such as \textit{\textbf{respiration rate}} and \textit{\textbf{skin conductance}}, offers alternative markers for interpersonal engagement \cite{darzi2021automated,  pecoraro2022psychophysiological,  tamantini2024fuzzy}. In RHHI, these techniques can provide indicators of motivation and engagement {and could potentially be used to, e.g., dynamically adapt the difficulty of a competitive or collaborative exercise for two users with different impairment levels \cite{darzi2021automated}.}

\end{tcolorbox}
\end{highlightbox*}

\subsection{Upper-Limb Training}
\label{sec:app:UpperLimb}
 
Several studies have investigated RHHI during upper-limb training, showing that healthy individuals track movement trajectories more accurately when haptically connected with a partner compared to training alone. These improvements have been linked to estimation of the partner's movement strategy to refine their own movement control~\cite{takagi2017physically,Takagi2018HapticInteraction,Takagi2019IndividualsGoal}.

Despite promising findings, few existing RHHI studies have evaluated these approaches in individuals with neurological impairments. Some studies have involved patient populations performing social rehabilitation games~\cite{baur2023competitive, Gorsic2017AGame}, however, the infrastructure used in these studies did not allow users to physically interact. Both studies emphasized social interaction to boost engagement during training, highlighting the importance of the interaction scenario in future RHHI work. The only notable exception is the study by Waters \textit{et al.} \cite{waters2024theradyad}. 
This work presented preliminary results on three chronic stroke participants paired with healthy partners during a trajectory tracking task, showing improved individual motor learning as a result of haptically interacting with an unimpaired user and thus validating the feasibility of such an approach for upper-limb training. {Future work could extend these RHHI findings during trajectory tracking tasks to larger cohorts in longitudinal studies to determine the effects of such training on sustained functional benefits.}

One promising research direction has been suggested by early studies on human–human interaction during handwriting~\cite{buscaglione2024human, way2024design}. To date, RHHI research has primarily focused on arm-level tasks, with limited attention to hand-specific activities requiring dexterous manipulation. Although various hand rehabilitation robots have been developed~\cite{ratz2024enhancing, ranzani2023design, kottink2024therapy}, none currently support direct haptic interaction between users. Enabling such interaction could significantly expand the range of functional and dexterity-based tasks that patients can perform in collaboration with therapists.

\subsection{Sit-to-Stand Training}
\label{sec:app:sitToStand}

The ability to transition from sitting to standing (StS) is fundamental for independent living. It is one of the most common functional tasks and correlates with clinical balance scores in stroke patients \cite{cheng1998sit}. Post-stroke individuals often struggle with StS due to deficits in strength, motor control, and proprioception, leading to asymmetric force distribution, slower movements, and greater pressure displacement \cite{boukadida2015determinants}.

Vianello \textit{et al.}~\cite{vianello2021human} introduced a dual-exoskeleton system  
that allows a therapist and patient to interact physically across multiple joints and contact points. This setup enables the therapist to guide the patient while incorporating programmable support features, such as balance assistance and safety measures essential for StS tasks. The system was tested with healthy individuals performing synchronized StS movements using shared visual feedback. To reduce fatigue during prolonged sessions, the exoskeletons offload part of the patient’s body weight, allowing them to focus on coordination rather than physical exertion. The study also explored two interaction configurations (Sec. \ref{sec:int_config}): (1) {joint-level coupling}, where a spring-damper system links the joint configurations of both users, and (2) {task-space coupling}, where a virtual connection between their centers of mass allows for flexible posture while preserving overall coordination.

By combining robotic assistance with therapist guidance, RHHI could provide a more adaptive and effective approach to StS training, and could be beneficially extended to standing balance training using a similar setup.

\subsection{Gait Training}
\label{sec:app:gait}

Conventional gait therapy typically involves therapists manually guiding patients to improve balance and gait patterns, often using partial body-weight support. During treadmill-based training, the therapist usually sits in front of the patient and manually assists leg movements. In overground training, body-weight support is also commonly employed, and multiple therapists may be required to support and guide both legs \cite{hesse2000mechanized}. Conventional gait therapy may include tasks such as stair climbing and obstacle avoidance; however, these require specialized body-weight support systems to safely assist patients during more complex maneuvers: equipment that is not available in all clinical settings.
Robot-Assisted Gait Training (RAGT) supports patients during high-intensity training \cite{gassert2018rehabilitation, ditommaso2023biomechanics}, but it typically facilitates only a limited range of exercises---most often treadmill walking, and less frequently, overground walking.

RHHI offers adaptive, interactive exercises tailored to individual patient needs and the possibility to extend RAGT to a wider range of exercises without prior trajectory definition.
Building on this concept, a recent framework virtually physically coupled two healthy participants wearing lower-limb exoskeletons and evaluated how different {interaction medium} affected inter-user coordination during treadmill walking \cite{kuccuktabak2023virtual}. The same framework was subsequently applied in a physical therapy setting with post-stroke individuals, where the therapist and patient were {bidirectionally} connected at the hips and knees \cite{kucuktabak2025therapist}. This system allows both sides of the exoskeleton to be actively engaged, enabling natural coordination and force exchange between the therapist and patient during gait training. Moreover, the virtual coupling parameters can be modified to accommodate different therapeutic goals, movement asymmetries, and patient abilities, offering flexibility across training scenarios. When compared with conventional manual therapy, this approach produced superior outcomes, including greater joint range of motion, longer steps, and higher muscle activation, alongside a high level of patient engagement and motivation.

{While these results are promising, one limitation of this approach is the need for multiple exoskeletons, which may restrict its practicality and scalability in clinical settings. To address such constraints, alternative designs have explored simplified or asymmetric coupling methods that reduce hardware complexity while maintaining interactive guidance.} Amato \textit{et al.} \cite{amato2024unidirectional} proposed and validated a system in which a therapist wearing an Inertial Measurement Unit (IMU)-based suit provides {unidirectional} haptic feedback to a patient in an exoskeleton during treadmill walking and obstacle avoidance.  Koh \textit{et al.} \cite{koh2021exploiting} implemented a {cross-limb interaction} by virtually connecting a robotic manipulandum to a robotic arm physically attached to the paretic leg of post-stroke patients. 
This setup enables {bidirectional} interaction, allowing therapists to actively influence the patient's leg movements and demonstrating improved kinematics during training.

Both approaches \cite{amato2024unidirectional, koh2021exploiting} support collaborative interaction and assist patients in forming internal movement models through therapist-guided feedback. While only Amato \textit{et al.} demonstrated obstacle avoidance, both systems have the potential to support a wider range of exercises. The main limitation lies in hardware constraints; for example, Koh \textit{et al.} use a grounded manipulator to assist the paretic leg, which may restrict its use in overground activities such as walking, stairs, or obstacles. Future developments could explore mobile manipulators to extend support beyond treadmill-based training (see Sec. \ref{sec:KeyConsideration} for future technical challenges). A key distinction between the two examples proposed is that the method by Amato \textit{et al.} uses a unidirectional connection---only the patient perceives haptic feedback---which lowers cost and setup time \cite{amato2024unidirectional}  (see Sec.\ref{sec:open_research_questions}). Another difference is that the approach of Amato \textit{et al.} supports interaction with both patient limbs, while Koh \textit{et al.} focuses on a single limb. Koh \textit{et al.} tested their system with hemiparetic stroke patients, targeting the impaired limb, but adapting the method for other populations, such as those with SCI, would require further development \cite{koh2021exploiting}.

\subsection{Group Therapy}
\label{sec:app:group}

Group therapy is widely used in rehabilitation, promoting social interaction and psychological support. Patients benefit by collaborating on exercises, sharing challenges, and engaging in motivational or competitive activities \cite{zanca2013group, church2019effectiveness}. 
However, group therapy is often limited to seated exercises, as therapists cannot provide individualized physical assistance simultaneously to multiple patients for functional movements such as walking or StS transitions. While seated exercises may be suitable for upper-limb rehabilitation, clinical guidelines emphasize the need for high-intensity gait training to improve locomotor function~\cite{hornby2020clinical}. 

RHHI could enable multiple patients to safely perform high-intensity functional exercises while engaging with therapists and peers, helping them achieve the necessary training dosage for recovery. However, to the best of our knowledge, no study has virtually connected multiple patients and/or therapists using robotic devices. Aprile \textit{et al.}  \cite{aprile2019improving}  
explored the feasibility of a robot-assisted rehabilitation area where a single therapist could supervise up to four patients simultaneously. However, the robotic devices were not virtually connected, limiting interactive engagement. Therapists primarily assumed a supervisory role rather than actively interacting with patients, which reduced their ability to assess performance, provide real-time feedback, and adapt interventions to individual needs.

Another potential group therapy application is music-based rehabilitation, where therapists and patients play instruments together. Michalko \textit{et al.}~\cite{michalko2024enhancing} proposed using RHHI to mediate interactions between expert and novice musicians, providing real-time feedback for posture and finger positioning. This concept could be extended to NR, engaging multiple patients in complex coordinated functional tasks that enhance synchronization and collaboration.
The benefits of music therapy go beyond motor synchronization---music itself can enhance emotional engagement, stimulate cognitive processes, and promote neuroplasticity, making it a powerful and motivating tool in rehabilitation contexts \cite{sihvonen2017music, leonardi2018role}.

RHHI has the potential to optimize therapist time while actively engaging multiple patients in interactive rehabilitation. Therapists can simultaneously monitor and adjust interventions, providing personalized guidance through robotic systems \cite{miller2022automated, adhikari2023learning}.
By adapting exercises to individual needs within a group setting, RHHI can enhance motivation, ensure appropriate challenge levels, and improve functional outcomes \cite{Baur2018TrendsGames}. Further discussion of research directions for this framework is provided in Sec. \ref{sec:open_research_questions}.

\subsection{Remote Therapy}
\label{sec:app:remote}

Telerehabilitation (rehabilitation delivered remotely) has proven effective in improving motor function, cognition, balance, speech, and communication skills \cite{nikolaev2022recent}. Studies have shown that telerehabilitation can yield outcomes comparable, or even superior, to conventional rehabilitation \cite{pitliya2025telerehabilitation}. A key advantage of telerehabilitation is the ability to provide rehabilitation to a broad population, particularly benefiting patients in remote areas with limited access to specialized care. However, some studies report higher satisfaction and motivation in clinic-based therapy due to increased social interaction \cite{cramer2019efficacy}. To mitigate this, approaches such as online therapist-led training \cite{smith2020combined} and multi-user interactions \cite{thielbar2020home} have been proposed. Videoconferencing and family has also been found to enhance motivation and clinical outcomes by fostering accountability in patients \cite{chen2020qualitative}. {RHHI can further support rehabilitation when physical proximity is limited, such as for patients requiring social distancing or individuals with Autism. In these cases, robots can act as mediators that facilitate engagement and communication between the patient and therapist, promoting social interaction, emotional expression, and even motor training \cite{santos2021design, fachantidis2020tauhe, jouaiti2019robot}.}

{RHHI can overcome telerehabilitation physical distance constraints \cite{nuara2022telerehabilitation}, enhancing engagement by facilitating real-time therapist-patient physical interaction in remote settings (e.g., therapist in the clinic, patient at home).} Therapists can initiate remote training sessions with a real-time assessment of the patient's performance through remote haptic interaction paired with videoconferencing 
\cite{kim2020remote}. 
{Alternatively, patients could connect to each other or to remote family members over the Internet and perform competitive, collaborative, or cooperative exercises together while tele/videoconferencing with each other, similarly to how modern videogames offer online play \cite{thielbar2020home}.}
Notably, since January 2024, Medicare insurance has supported the cost of personal exoskeletons, paving the way for RHHI integration into mainstream care \cite{ExoMedicaid}, allowing continuous rehabilitation support even after discharge.

\section{Key Considerations for RHHI in NR}
\label{sec:KeyConsideration}

This section explores key advancements necessary to further expand the application of RHHI in rehabilitation. We first discuss the design and control of robotic devices, followed by the networking between paired robots. Finally, we address essential factors for improving accessibility and widespread adoption of these technologies.

\subsection{Robot Design}
\label{sec:KC:robot_design}

\textit{RHHI for upper-limb NR should evolve beyond grounded systems to enable activities of daily living, including dexterous interactions}.
Upper-limb RHHI has been explored using both single- and multi-DoF robots. However, most studies focus on grounded robotic systems for seated tasks. Expanding RHHI to mobile robots or portable upper-limb exoskeletons could broaden applications, allowing patients to perform functional activities while engaging in dynamic movements.

To date, no work has explored hand exoskeletons to establish haptic connections between individuals. Dexterous manipulation tasks, with their complexity and variability, are well-suited for RHHI, enabling users to collaboratively develop manipulation strategies. Incorporating sensory feedback, such as haptic cues about the manipulated object, could enhance interaction and promote embodiment 
 \cite{pinardi2023impact, noccaro4884432vibrotactile}.  

\textit{RHHI for lower-limb NR should move toward multi-DoF, full-body interactions to better support complex gait and balance tasks}. Recent advances have expanded RHHI to lower-limb applications, ranging from single-DoF ankle movements to more complex gait scenarios using exoskeletons or robotic manipulators. However, current implementations are limited by technological constraints. Preliminary studies on ankle tracking \cite{short2023haptic} and dyadic exoskeleton therapy \cite{vianello2023exoskeleton, kuccuktabak2023virtual} have focused exclusively on sagittal plane actuation, neglecting essential coronal and transverse plane movements utilized in human motion. 
One approach to address this limitation involves end-effector robots \cite{koh2021exploiting} to mediate lower-limb interactions between therapists and patients. Alternatively, cable-based soft exoskeletons, \cite{basla2024enhancing} could facilitate dynamic interactions, though this remains largely unexplored. While both strategies improve joint alignment and synchronized movement, they may not provide sufficient power or multi-DoF assistance for more severely impaired patients. Future research should prioritize multi-axis RHHI capable of enabling comprehensive lower-limb interactions between users.

\textit{RHHI must address cost and accessibility challenges without compromising therapeutic effectiveness}. A major barrier to the widespread adoption of RHHI in NR is the need for multiple robotic devices, making the infrastructure prohibitively expensive for many clinics and hospitals. This financial constraint limits accessibility and practical implementation in routine therapy. To mitigate costs, Amato \textit{et al.} \cite{amato2024unidirectional} proposed IMU sensors to replace robotic devices, reducing potential costs at the expense of limiting haptic feedback. Similarly, Koh \textit{et al.} \cite{koh2021exploiting} introduced joystick-based control, which reduces costs but can diminish the sense of embodiment. Although these approaches offer more affordable options, they come with trade-offs that could impact therapeutic effectiveness.
Future research should evaluate whether these cost-saving measures adequately compensate for the loss of haptic interaction and embodiment, or if alternative approaches can be developed to balance affordability with functional benefits. {At the same time, design under co-creation schemes involving clinicians and engineers ensures patient engagement and adherence to therapies \cite{dobe2023co}.}

\begin{highlightbox}
\refstepcounter{mybox}
\begin{tcolorbox}[colback=blue!5!white, colframe=blue!75!black, title= Box 2: Learning by Demonstration and RHHI, 
fonttitle=\bfseries, boxsep=2pt,    
  left=4pt,       
  right=4pt,      
  top=2pt,        
  bottom=2pt      
  ]
\label{box:lbd}

\textbf{What is it?} Learning by Demonstration (LbD) models therapist strategies from dyadic interactions, enabling adaptive controllers that improve robotic assistance and patient outcomes \cite{mulian2024mimicking}.

\textbf{Why LbD and RHHI?} (1) Captures therapist-patient interaction forces during physical guidance; (2)  Enhances therapist trust and promotes patient-specific adaptation;
(3) Enables robots to learn assistance strategies without needing hard-coded rules and improve over time through ongoing feedback and demonstration.

\textbf{Recent examples: } (1) \textit{Upper-limb exoskeleton} - Luciani et al. \cite{luciani2024imitation, luciani2024therapists} introduced an LbD framework that captures therapist interaction forces while guiding a patient’s movement using an upper-limb exoskeleton. 
(2) \textit{Lower-limb exoskeleton} - Sankar et al. \cite{sankar2024action} demonstrated preliminary results in this direction, using an LbD model trained with data from a therapist-patient dyad connected to lower-limb exoskeletons. 

\textbf{What is next?} 
Incorporating additional sensor data (e.g., heart rate, skin conductance) could further refine control policies. These adaptive systems could improve over time through ongoing therapist feedback and demonstrations.

\textbf{Takeaway: }Applying LbD to RHHI can offer insights into therapist-patient interactions, enabling the design of models that learn personalized assistive or resistive strategies to the patient's needs. 


\end{tcolorbox}
\end{highlightbox}

\subsection{Robot Control}
\label{sec:KC:robot_control}

\textit{Accurate haptic rendering through robot transparency is a fundamental requirement for effective RHHI} \cite{kuccuktabak2023haptic}. Transparency ensures that the forces and motions experienced by the user reflect the intended interaction---minimizing the forces perceived due to the robot’s own dynamics---making it a prerequisite for delivering precise, natural, and therapeutically meaningful feedback \cite{lawrence1993stability,colgate1997passivity}. This is especially critical in NR contexts, where the fidelity of physical interaction directly affects motor learning and recovery outcomes \cite{proietti2016upper, jarrasse2014perspective}.

Despite its importance, achieving transparency remains challenging. Most rehabilitation robots are designed for robustness and high output forces, which inherently introduces unwanted dynamics such as friction, inertia, and gravitational loads \cite{hogan1985impedance, just2016motor, just2017feedforward}. These effects can mask or alter the forces exchanged between human partners and robots, thereby reducing the clarity and utility of haptic cues \cite{noccaro2023evaluation, just2018exoskeleton}. Feedforward compensation, often used to eliminate known dynamics such as gravity and friction, has been widely adopted \cite{ jarrasse2011connecting, proietti2016upper}, but modeling inaccuracies and user variability often lead to residual errors that reduce transparency and consistency across users.

RHHI effectiveness hinges not only on accurate force rendering, but also on intuitive role assignment and shared control between a user and their robot \cite{jarrasse2014perspective, jarrasse2014slaves, proietti2016upper}. In this framework, the robot should function as a seamless facilitator, supporting natural, bidirectional exchange and allowing therapists to transfer their expertise directly through physical interaction. One concrete example of this is adaptive arm weight compensation \cite{jarrasse2011connecting, just2020human}, which virtually unloads the limb to allow effortless movement while preserving haptic cues, thus enhancing both comfort and value of perceiving haptic information. Ultimately, the challenge in RHHI control lies in delivering haptic rendering that is not only technically precise, but also contextually meaningful, transforming the robot from a source of interference into a channel for physical communication \cite{reinkensmeyer2004robotics, krebs1998robot, ajoudani2018progress}.

\textit{Ensuring user safety during RHHI is equally essential}. While RHHI increases adaptability in therapy, it also introduces unpredictability---for example, unintentional therapist movements or patient reflexes. To mitigate these risks, Amato \textit{et al.} \cite{amato2024unidirectional} proposed filtering high-frequency components of the interaction to maintain controlled and stable coupling. Hierarchical constraint-based controllers have also been developed to prioritize safety over rendering performance when necessary \cite{kuccuktabak2024physical}. Moreover, recent advances in real-time perception have enabled safe interaction with dynamic and deformable human body surfaces \cite{Sommerhalder}. Intelligent systems based on machine learning and advanced collision detection can further improve safety by distinguishing intentional actions from unintended contact or perturbations \cite{haddadin2017robot, deluca2006collision, haddadin2017robotcollisions}.

\subsection{Device Networking}
\label{sec:KC:Networking}

Most current RHHI systems rely on Ethernet and local connections, requiring devices to be grounded to a shared router. While effective for research environments, this limits system mobility and flexibility. Transitioning to WiFi or remote connections in clinical settings introduces several challenges, particularly concerning data security and performance.

\textit{Ensuring patient data privacy is paramount}, especially in healthcare settings where sensitive information is routinely processed. Robust encryption methods \cite{deng2022security} are essential to protect data during transmission and prevent unauthorized access. Furthermore, strong authentication protocols must verify users' identities, ensuring that only authorized personnel can access patient information. Compliance with regulatory frameworks \cite{maddali2024ai} is also necessary to maintain high standards of data protection and confidentiality.

\textit{Unstable networks like WiFi can disrupt RHHI by causing delays, underscoring the need for reliable connectivity and advanced control}. Switching to less stable networks introduces potential delays and synchronization issues, which are critical for effective bidirectional interaction in RHHI. As observed by Kim \textit{et al.} \cite{kim2020remote}, delays up to 500 ms were manageable for slow, controlled movements but negatively impacted fast, dynamic tasks. In fact, their communication protocol with checksum verification maintained safety for slow movements but was inadequate for fast tasks such as spasticity assessments. Similarly, Ivanova \textit{et al.} \cite{ivanova2021short} found that while humans can adapt to delays up to 60 ms, longer delays degrade interaction quality in the context of RHHI.

To improve performance, future research should focus on reducing latency through optimized data compression, faster communication protocols, and prioritization of critical data packets. Predictive algorithms, such as those explored by Penco \textit{et al.} \cite{penco2023prescient}, can compensate for network delays by adjusting movement timing to maintain synchronization. Additionally, integrating local processing capabilities into RHHI devices can enable critical functions to be managed independently of network conditions, minimizing the impact of communication delays on performance.

\subsection{Translation of RHHI into Clinical Care}
\label{sec:KC:translationToClinic}


\textit{Structured and realistic training methods are essential for preparing therapists to adopt RHHI, ensuring both confidence and competence in clinical use}. Studies have shown that increased training improves trust and acceptance of new rehabilitation technologies \cite{hasson2023neurorehabilitation, celian2021day, mortenson2022therapists}. This is particularly important in RHHI, where therapists are not only facilitators but also end-users alongside patients.
For this purpose, therapist-standardized procedures should be implemented. For instance, in Kucuktabak \textit{et al.} \cite{kucuktabak2025therapist}, therapists participated in training sessions where they were haptically connected to healthy team members simulating asymmetric gait before working with patients. An alternative solution is to artificially induce impairments in healthy participants to simulate neuromotor deficits. Manczurowsky \textit{et al.} \cite{manczurowsky2024induced} introduced "dysfunctional electrical stimulation" to involuntarily activate the hamstrings of healthy participants during walking, mimicking neuromotor impairments. This method could improve therapist training by exposing them to realistic patient challenges in a controlled environment \cite{buddeoutside}.
Hasson \textit{et al.} \cite{hasson2019learning, hasson2019visual} developed a system where participants modified a virtual stroke survivor’s gait through robotic manipulanda. Simulated training methods provide therapists with unlimited exposure to patient movement dynamics, reducing risks associated with real patient scenarios. Additionally, impairment simulations using visual noise or mechanical constraints \cite{noccaro2022visual,buscaglione2023assessment} can further enhance training realism.

\textit{RHHI should support a broad and functional range of exercises through standardized yet customizable task libraries to meet diverse NR needs}. 
{A key advantage of RHHI over conventional robotic rehabilitation lies in its ability to support a broader range of exercises. However, most existing RHHI studies for both upper and lower limbs focus on stereotyped reaching or tracking tasks \cite{kuccuktabak2021human}, which have limited functional relevance. Future developments should incorporate more naturalistic and multi-joint functional tasks, e.g., object manipulation, and obstacle negotiation \cite{celian2021day,amato2024unidirectional}. Establishing a standardized yet customizable exercise library, developed in collaboration with therapists, would balance consistency and adaptability to effectively address diverse rehabilitation goals.
In Box. \ref{box:roadmap}, a Road Map to Clinical Adoption is summarized.}

\begin{highlightbox*}
\refstepcounter{mybox}
\begin{tcolorbox}[colback=blue!5!white, colframe=blue!75!black, title= Box 3: A Road Map to Clinical Adoption, 
fonttitle=\bfseries, boxsep=2pt,    
  left=4pt,       
  right=4pt,      
  top=2pt,        
  bottom=2pt      
  ]
\label{box:roadmap}

\textbf{Co-Creation and Stakeholder Involvement - } 
{The success of robot-mediated Human–Human Interaction (RHHI) technologies relies on co-creation among patients, clinicians, engineers, and caregivers to ensure clinical relevance, usability, and robustness. Therapists should be engaged early through workshops and iterative design feedback to align features with therapeutic needs \cite{nicora2025healthcare, aprile2025rehabilitation}. 
Group discussions with therapists are valuable for identifying barriers and expectations, as shown in prior Technology Acceptance Model studies \cite{luciani2023technology}.
Effective clinical translation of RHHI requires coordinated efforts across engineering, clinical, regulatory, and institutional domains—addressing training, certification, workflow alignment, and safety—while relying on therapists as key mediators who integrate these technologies into practice and bridge engineering design, patient needs, and healthcare delivery \cite{thawisuk2025perspectives, huang2018translation}.}

\textbf{Adaptation and Clinical Integration - }
{Effective clinical adoption of RHHI requires structured training and defined competency pathways, including didactic instruction, supervised practice, standardized evaluations, and periodic re-certification. Integration into existing workflows should be supported by pilot studies assessing feasibility, therapist acceptance, and patient safety, with clearly delineated staff roles and safety protocols. Embedding RHHI within established frameworks such as the International Classification of Functioning, Disability and Health, and aligning it with clinical assessment scales will facilitate implementation. Practical considerations are critical: therapists value minimal setup time, reliable operation, and compatibility with typical 45–60-minute sessions.}

\textbf{Legal and Regulatory Considerations - } {Compliance with regulatory and ethical standards (HIPAA, GDPR, IEC/ISO, FDA/MDR) is essential, alongside clear attention to liability, cost, and maintenance. Embedding safety, usability, and risk management from the design stage—supported by post-market monitoring—ensures RHHI systems are reliable, accepted, and ready for routine clinical use\cite{ali2025ethical}.} 

\end{tcolorbox}
\end{highlightbox*}

\section{Open research directions}
\label{sec:open_research_questions}

In this section, we outline key open research questions in the field of RHHI that will help identify the impact of this technology on NR, the factors that contribute to its effectiveness, and maximize its potential for clinical implementation.

\subsection{
What are the foundational mechanisms behind {motor learning} in RHHI in NR?}
\label{sec:RQ_motor_learning}

Numerous studies on RHHI in healthy individuals have explored gains in individual performance (i.e., motor learning) following haptic training with a partner \cite{waters2024motor}. These studies often use trajectory tracking tasks, sometimes incorporating perturbations (e.g., force fields, visuomotor rotations) to examine changes in task performance. Some findings suggest that haptic training with a partner accelerates motor learning compared to training alone \cite{Ganesh2014, Batson2020} or compared to training with a direct connection to the target trajectory \cite{ivanova2022interaction}. While gains have been attributed to mutual information sharing through the haptic connection \cite{takagi2017physically, Takagi2018HapticInteraction}, these learning benefits are not consistently observed across all studies.

Research has found no significant differences in individual learning outcomes between haptically connected and unconnected training in certain contexts, particularly during multi-joint reaching tasks \cite{Beckers2020} and lower-limb ankle tracking tasks \cite{kim2022effect}. These findings suggest that, in such cases, improvements may result more from the mechanics of interaction---such as error averaging across users---than from mutual adaptation of movement strategies, which is more commonly observed in upper-limb studies \cite{short2023haptic}. These heterogeneous results highlight that the learning effects of RHHI are likely influenced by multiple factors, including robotic system characteristics (e.g., quality of haptic rendering), task-specific constraints (e.g., DoFs, workspace size, and difficulty), and the limbs involved (e.g. upper or lower ones). 

Studies in healthy individuals have often aimed to isolate the effects of haptic interaction on motor learning by ensuring participants were unaware they were virtually connected to a partner. These experiments excluded visual and verbal communications to minimize cognitive adaptations associated with explicit coordination. In contrast, for NR applications, explicit awareness of the interaction cannot be neglected since it is crucial for both therapists and patients. Conventional manual therapy relies on multiple channels---visual observation, physical guidance for assistance or resistance, and verbal instruction---to direct and adapt patient movements. 

Given these differences, the mechanisms of motor learning in RHHI for NR are likely to be more complex than those observed in controlled laboratory settings with healthy individuals
\cite{takagi2017physically, Takagi2018HapticInteraction}. Unlike experimental studies that isolate haptic input, clinical applications will involve the combined influence of multisensory feedback and cognitive engagement. 
{For example, a recent study that tried to combine haptic interaction with engaging visual feedback failed to replicate previously observed motor learning benefits as a result of haptic interaction between partners \cite{hossain2025combining}.}
While prior work provides a useful foundation, further research is needed to characterize how learning unfolds when patients and therapists interact through RHHI with full awareness of the physical and social dynamics involved.

\subsection{
{What is the role of motivation, engagement, and gamification in RHHI?}}
\label{sec:RQ_engagement}

Motivation plays a crucial role in NR, and RHHI has been recognized for its potential to enhance motivation. Increased motivation may lead to longer exercise durations and greater intensity, which in turn could improve functional outcomes. Several studies on RHHI have reported motivational benefits compared to exercising alone, likely due to the social interaction component \cite{GajadharDoubleFun,Gorsic2017CompetitiveIntensity,Baur2018RobotSupportedTraining,baur2023competitive,Mace2017BalancingTraining}. 

Some research has further shown that competitive RHHI games can encourage longer and more active participation \cite{Gorsic2017AGame, baur2023competitive}, reinforcing the idea that social interaction enhances engagement. Additionally, Goršič \textit{et al.} \cite{Gorsic2020EffectsGame} found that replacing a human opponent with a human-like computer-controlled opponent resulted in lower motivation levels, suggesting that RHHI benefits may not be easily replicated by artificial agents \cite{takagi2017physically}. However, gamification effects are not necessarily additive, as studies on virtual rehabilitation environments indicate that multiple gamification elements do not always enhance engagement \cite{laver2017virtual}.

Studying motivation in RHHI is challenging because gamification and social interaction are often examined separately from other factors (as discussed in Sec. \ref{sec:RQ_motor_learning}). For instance, many gamification studies typically involve contactless games such as Pong or air hockey \cite{Gorsic2017CompetitiveIntensity,Novak2014IncreasingGameplay,Gorsic2017AGame,Pereira2019ImpactInvolvement,baur2023competitive}.
Additionally, social interaction and gamification may unintentionally alter task dynamics. For instance, human opponents behave more unpredictably than computer-controlled opponents, affecting motion dynamics. Similarly, if participants receive different interaction instructions
\cite{takagi2023competition}, it may impact both social engagement and movement execution, making it difficult to isolate the specific mechanisms underlying changes in effort or movement quality.

While individual preferences vary, interpersonal differences complicate gamification research in RHHI. Preferences for competition, cooperation, visual complexity, and difficulty vary widely across individuals, limiting generalizability. Player profiling can help, but such variability hinders broad conclusions. Still, gamification offers a cost-effective way to boost motivation in RHHI without added hardware, though the links between engagement and learning outcomes remain unclear. Therefore, future work should identify effective game designs for RHHI and clarify how task dynamics and user preferences shape rehabilitation results.

\subsection{
Who should be the partner in RHHI, and should they interact as opponents or teammates?}
\label{sec:RQ_social_interaction}

Patients in RHHI can interact with three potential groups of people: therapists, other patients, and healthy loved ones (e.g., spouses or friends). The most appropriate pairing depends on the interaction’s purpose, such as clinical assessment, treatment, or social engagement. For example,  a clinical assessment scenario, as discussed in Baur \textit{et al.} \cite{baur2019beam}, requires a therapist as the partner. 
However, therapists may not be ideal partners for competitive games, as patients might not perceive them as fair opponents or could find clinical feedback distracting \cite{Gorsic2017CompetitiveIntensity}. {Although matching patients with others of similar ability may appear ideal for competitive settings, studies indicate that intelligent difficulty adaptation algorithms can enable individuals to engage in competition with unimpaired loved ones despite differences in motor ability} \cite{Gorsic2017AGame}. {Such interactions may offer additional motivational benefits by allowing patients to exercise with familiar and emotionally supportive partners.} In fact, studies suggest that participants have a more positive experience when they ``get along" with their partner \cite{Novak2014IncreasingGameplay, Gorsic2017CompetitiveIntensity, pereira2021use}. 
Beyond partner type, individual personality traits and cognitive abilities must also be considered \cite{pereira2021use}.

On the other hand, depending on partner type, interaction dynamics may need to be adjusted. As discussed in Sec. \ref{sec:taxonomy}, a virtual spring with adjustable stiffness can connect joint positions between users. While this method is typically used to link a patient with a therapist, some research suggests that novices (i.e., patients) may learn better when paired with other novices rather than experts \cite{ganesh2014two}. Alternatively, if a virtual spring is activated intermittently, e.g., through an on/off button, even an unimpaired loved one could assist by selectively providing support at critical moments \cite{Baur2018RobotSupportedTraining}.

The interaction between participant characteristics and RHHI dynamics likely affects engagement and outcomes in complex ways. Some individuals may prefer exercising alone over collaborating with an unsuitable partner. Furthermore, partner selection is often limited by availability. In inpatient settings, unimpaired loved ones may only be available during visits and might prefer to socialize rather than exercise, leaving patients to work with therapists or other patients. Given these constraints, patients may still have preferences about their partners, which should be considered in rehabilitation plans.

Similarly, in home-based rehabilitation, partner choice depends on the patient’s level of impairment. Severely impaired individuals may rely on caregivers or family members, whereas more mobile patients may seek out preferred partners. Technology could further expand these options by enabling patients to connect with suitable partners online (Sec. \ref{sec:app:remote}). Future research should explore how partner dynamics influence RHHI outcomes, while considering both availability and patient autonomy.

\subsection{
How much do therapists rely on haptic feedback in RHHI?}
\label{sec:RQ_haptic_feedback_therapist}

Two primary interaction approaches in RHHI are discussed in the literature (Sec. \ref{sec:taxonomy}): unidirectional and bidirectional. In the unidirectional approach \cite{amato2024unidirectional, short2025effects, Ganesh2014}, one user receives haptic feedback from their partner, while the partner does not perceive any interaction forces. This can be implemented by using IMU sensors to track movement \cite{amato2024unidirectional}, by recording and replaying executed trajectories \cite{Ganesh2014, short2025effects}, or by setting a zero-stiffness spring in one robot. In this scenario, one user---likely the therapist---must rely solely on visual and auditory feedback for task execution.

In bidirectional interaction, both users receive haptic feedback while interfacing with separate robotic devices \cite{koh2021exploiting}. This enables a therapist to perceive their patient’s movements through force feedback, adding an additional communication channel beyond visual and auditory cues. While the unidirectional approach reduces costs by requiring only one robot, its effectiveness compared to bidirectional interaction remains unclear. Available studies have shown that unidirectional and bidirectional interaction during wrist and ankle trajectory tracking produce similar performance improvements for the less-skilled partner in pairs of healthy individuals \cite{short2025effects}. However, it remains uncertain whether these findings hold when a therapist interacts with a patient with sensorimotor impairments. In clinical settings, therapists may compensate for the lack of haptic feedback by visually observing the patient’s performance and adapting their movements accordingly.

A hybrid approach, where haptic feedback varies based on motor skill differences, may offer a promising solution. In standard bidirectional RHHI, the more-skilled users must exert greater effort to compensate for the deviations of their less-skilled partner. This imbalance can make interactions physically demanding for the therapist when connected to a patient. Noccaro \textit{et al.} \cite{noccaro2025robot} demonstrated that asymmetric connections with different stiffness for each partner can reduce the effort of the more-skilled user while balancing effort to match baseline conditions without haptic connection.

This asymmetric approach is particularly well-suited for NR, as it can reduce the physical effort required by the therapist while retaining the benefits of bidirectional haptic communication. An open question in RHHI is the extent to which therapists rely on physical interaction to make real-time corrections during NR. Understanding this reliance is crucial for designing systems that balance automation with the nuanced, hands-on adjustments therapists provide. Asymmetric connections may offer a way to preserve critical haptic information for the therapist while mitigating other potential drawbacks, ensuring effective and scalable RHHI solutions in clinical settings.
 
\subsection{
Can RHHI scale NR while preserving individualized care for group therapy applications?}
\label{sec:RQ_group_therapy}

Some studies have shown that a single therapist can supervise multiple patients by using robotic devices in a rehabilitation setting \cite{aprile2019improving, wuennemann2020dose, miller2022automated, adhikari2023learning}. For instance, Miller \textit{et al.} \cite{miller2022automated} proposed an automated robotic gym where patient-robot assignments dynamically adjust to optimize rehabilitation outcomes. While this study was conducted in simulation, several rehabilitation centers, such as Fondazione Don Carlo Gnocchi (Italy), Shirley Ryan AbilityLab (USA), and SRH Gesundheitszentrum Bad Wimpfen (Germany), have established robotic gyms equipped with multiple robots. These facilities have demonstrated outcomes comparable to traditional rehabilitation setups, with a single therapist effectively supervising two to four patients. However, none of these systems has yet implemented a networked RHHI framework that allows physical interaction between patients and therapists.

Integrating RHHI into robotic gyms could address several key challenges. Currently, therapists must divide their attention across multiple patients, increasing the risk of suboptimal therapy, patient fatigue, or, in worst-case scenarios, injuries if critical adjustments are missed. To mitigate these issues, therapists must gain a comprehensive understanding of each patient’s performance through multisensory channels, including visual observation, verbal communication, and haptic feedback. Additionally, centralized monitoring systems could aggregate patient data from all robots, reducing therapist workload and enhancing oversight. This framework could further benefit from dynamic adjustment systems \cite{miller2022automated, adhikari2023learning} which optimize patient-robot assignments and introduce patient-therapist or patient-patient coupling to maximize rehabilitation outcomes.

An exciting research direction involves modeling virtual interactions between more than two users. For example, a single therapist could be virtually connected to multiple patients, perceiving a combined force that reflects each patient’s movement execution. This force could be modulated by adjusting the stiffness profiles of the virtual springs connecting the therapist to each patient. If stiffness is modeled symmetrically (see Sec. \ref{sec:RQ_haptic_feedback_therapist} for details on therapist-specific stiffness profiles), the resulting force could be excessively high, increasing therapist effort. Therefore, stiffness modulation will likely be essential to balance force distribution while considering the individual abilities of each patient.

Another key question is the scalability of such interaction networks. Just as in a crowded room where overlapping voices create noise, there may be a limit to how many patients a therapist can engage with effectively in a NR session before interactions become too complex. Therefore, it remains unclear whether these results would hold in settings where patients and therapists engage in direct physical interaction. Determining the upper limit of networked patient-therapist interactions could help maximize rehabilitation efficiency while maintaining individualized patient care.

\subsection{
How will future rehabilitation based on RHHI reshape the role of {therapists and caregivers}?}
\label{sec:RQ_therapist_substitution}

The evolution of robotics in rehabilitation follows trends seen in other fields, such as industrial automation. In the early phases of robotics, a strict separation existed between robots and humans. Indeed, industrial robots performed repetitive tasks in isolation from human operators for safety reasons. However, this division is narrowing, and the long-term goal of fully autonomous robots is gradually shifting towards collaborative systems \cite{nardo2020evolution}.

Similarly, rehabilitation robotics is transitioning toward more interactive and adaptive systems, with RHHI enabling real-time therapist-patient interaction. Rather than replacing therapists, these technologies have the potential to enhance therapy outcomes while increasing the need for highly skilled professionals trained in advanced robotic-assisted rehabilitation. {The benefits of RHHI have been widely discussed in this paper; their impact on therapist employment and job roles remains an important topic to be investigated to foster education and collaboration within an interdisciplinary community \cite{motus_academy, polimi_rehabtech}.}

Prior studies have reported therapists' concerns about robots potentially replacing them in the future \cite{postol2023we}. However, RHHI is designed to complement therapist expertise rather than substitute it. In this framework, robotic systems and therapist knowledge evolve together, mutually enhancing each other’s capabilities. While algorithms provide automation, they lack the ability to fully capture the nuanced decision-making and adaptability required for effective rehabilitation.
{Beyond therapists, home rehabilitation with RHHI can actively involve caregivers by enabling them to guide patients through the robotic system. Currently, such guidance is often ad hoc and unstructured. RHHI could standardize this interaction, combining caregivers’ adaptability with the knowledge and capabilities of rehabilitation robots.}

The integration of clinical expertise within the robotic control loop remains a key area for further research. Over the next years, studies should actively engage rehabilitation stakeholders—therapists, patients, caregivers, clinicians, health service providers, and healthcare administrators—through qualitative research and feasibility testing within existing medical models. This will ensure that RHHI continues to enhance rehabilitation without compromising the essential human element of therapy.

\subsection{
Is the impact of technologies needed for RHHI high enough to justify additional costs?} 
\label{sec:RQ_costs}

To rigorously evaluate the potential of RHHI, researchers should conduct larger clinical trials with comparisons to conventional manual therapy. So far, studies comparing robotic-assisted rehabilitation to traditional approaches have not produced results strong enough to justify the cost of robotic devices. To address the financial barriers that have limited widespread adoption, companies are now introducing Hardware-as-a-Service models \cite{forbes2025}, which offer robotic devices through leasing programs. This shift could significantly improve accessibility for clinics and researchers, enabling broader clinical validation of RHHI. As part of this effort, robust Health Technology Assessment will be essential to evaluate the clinical and economic value of RHHI solutions \cite{banta2003development}.  
An important aspect to consider is the therapist’s willingness to adopt RHHI compared to conventional robotic approaches. The perception of greater control over the exercise and its impact on the patient may enhance perceived usability, potentially making RHHI a more effective and acceptable solution in clinical practice. To validate this hypothesis, it will be essential to include therapists in clinical trials through structured interviews and usability assessments---recognizing that in the RHHI framework, therapists are not merely facilitators but end-users alongside patients.


\section{Conclusions}
The evolution of rehabilitation robotics is moving toward interactive, patient-centered approaches. RHHI has the potential to revolutionize current scenarios and enhance clinical outcomes while preserving the essential human connection between patients and therapists. This position paper has outlined the key advantages of the RHHI approach and the research directions needed to effectively transfer current findings to clinical settings.  

We advocate for a RHHI framework where robotic technology is not a substitute for therapists but rather an augmentation of their expertise and capabilities. While robots offer strength, precision, and measurable outcomes, therapists remain vital for delivering personalized, multisensory care and nuanced decision-making.
Our multidisciplinary team---including clinicians, therapists, and engineers---has outlined key directions for future research.
First and foremost, future efforts should focus on the design of robotic systems and control algorithms that prioritize task functionality, safety, and cost-effectiveness. In parallel, further development is needed to enable robust networking between devices, with particular attention to privacy and system stability. Moreover, RHHI integration into clinical practice requires the involvement of therapists when designing system features and defining key functional exercises.
Fundamental research challenges were also identified and should be addressed to advance the field of RHHI. This requires deeper insight into motor learning mechanisms during RHHI, including motivation, engagement, partner characteristics, and interaction dynamics, supported by multidisciplinary research on social and physical interaction. 

RHHI represents a paradigm shift in rehabilitation, offering a powerful synergy between human and robotic complementary contributions. However, realizing its full potential requires a concerted effort from researchers, clinicians, and health care policymakers. By addressing these challenges, we can ensure that RHHI remains a tool for empowerment—enhancing, rather than replacing, the human connection that lies at the heart of rehabilitation.

\bibliographystyle{bibliografy/myIEEEtran}
\bibliography{references}

\end{document}